\documentclass[runningheads]{llncs}
\usepackage[T1]{fontenc}
\usepackage{graphicx}
\usepackage{amsfonts}
\usepackage{amsmath}
\usepackage{subcaption}
\usepackage{caption}
\usepackage{bm}
\usepackage[hidelinks]{hyperref}
\usepackage{cleveref}
\usepackage{url}
\usepackage{orcidlink}

\usepackage{color}

\begin{document}
%



\title{Verifying Machine Unlearning with \\ Explainable AI}

\titlerunning{Verifying Machine Unlearning with Explainable AI.}
%
\author{Àlex Pujol Vidal\inst{1,2} \orcidlink{0009-0004-0126-6692} \and
Anders S. Johansen\inst{1}\orcidlink{0000-0002-9330-522X} \and
Mohammad N. S. Jahromi\inst{1}\orcidlink{0000-0002-6332-7567} \and Sergio Escalera\inst{1,2}\orcidlink{0000-0003-0617-8873} \and  Kamal Nasrollahi\inst{1,3}\orcidlink{0000-0002-1953-0429} \and Thomas B. Moeslund\inst{1}\orcidlink{0000-0001-7584-5209}}
\authorrunning{A. Pujol et al.}
%
\institute{Aalborg University \and University of Barcelona \and Milestone Systems\\
\inst{1}\email{\{alexpv,asjo,mosa,kn,tbm\}@create.aau.dk} \hspace{5pt} \inst{2}\email{sescalera@ub.edu} \hspace{5pt} \inst{3}\email{kn@milestone.dk}}
\maketitle              
\begin{abstract}
We investigate the effectiveness of Explainable AI (XAI) in verifying Machine Unlearning (MU) within the context of harbor front monitoring, focusing on data privacy and regulatory compliance. 
With the increasing need to adhere to privacy legislation such as the General Data Protection Regulation (GDPR), traditional methods of retraining ML models for data deletions prove impractical due to their complexity and resource demands. MU offers a solution by enabling models to selectively forget specific 
learned patterns without full retraining.
We explore various removal techniques, including data relabeling, and model perturbation. Then, we leverage attribution-based XAI to discuss the effects of unlearning on model performance. 
Our proof-of-concept\footnote{The code and additional visualizations can be found at: \url{https://github.com/ASJAAU/Explaining_unlearning.git}} introduces feature importance as an innovative verification step for MU, expanding beyond traditional metrics and demonstrating techniques' ability to reduce reliance on undesired patterns. Additionally, we propose two novel XAI-based metrics, 
Heatmap Coverage (HC) and Attention Shift (AS), to evaluate the effectiveness of these methods.
This approach not only highlights how XAI can complement MU by providing effective verification, but also sets the stage for future research to enhance their joint integration.



\end{abstract}

\begin{figure}[h]
    \centering
    \includegraphics[width=\textwidth]{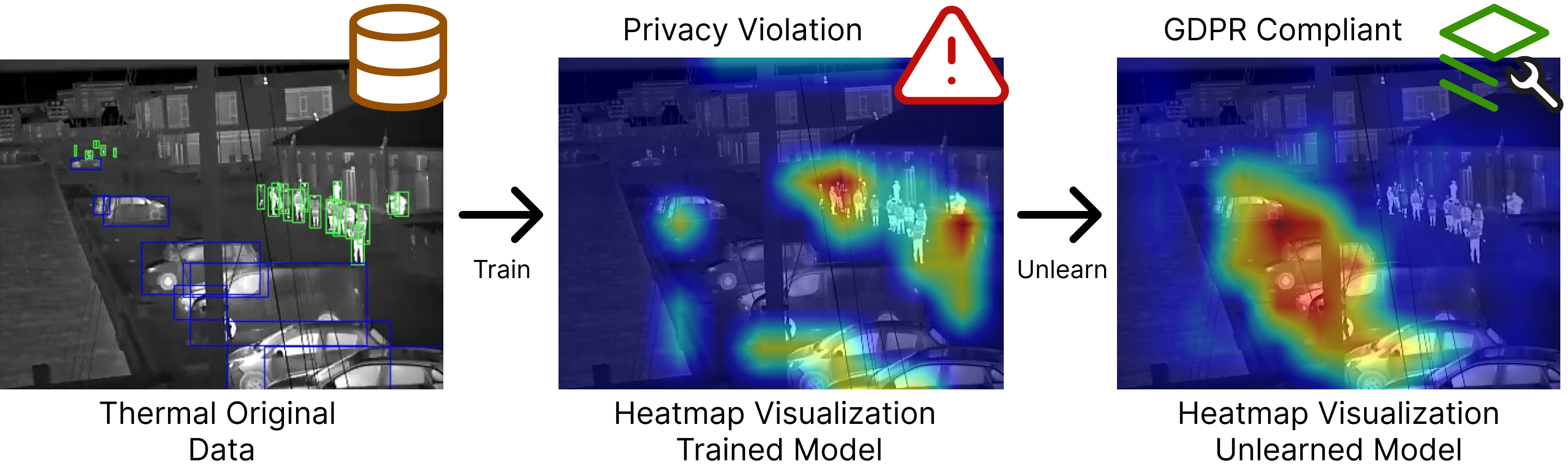}
    \caption{XAI and MU applied in the harbor front use-case, mitigating GDPR privacy violation. 
    XAI heatmaps show that the trained model captures human patterns while the unlearned model ignores them.}
    \label{fig:motivation}
\end{figure}
\section{Introduction}

In the rapidly evolving landscape of Artificial Intelligence (AI) and Machine Learning (ML), adapting to new data while ensuring privacy and regulatory compliance is crucial~\cite{xu2024machine}.
The EU's "right to be forgotten" highlights the need to modify or remove ML predictions without degrading performance ~\cite{nguyen2022survey,triantafillou2024makingprogressunlearningfindings}.
In domains such as security, healthcare, and activity monitoring addressing privacy concerns is vital for user trust and legal compliance. 
These fields often handle sensitive data, making it essential to enable data removal on request without disrupting ML systems ~\cite{shaik2023exploring}.
For instance, harbor front monitoring involves continuous analysis of potentially sensitive data. The General Data Protection Regulation (GDPR) imposes strict regulations on processing sensitive data, including any data that can identify individuals, such as images containing people. Consequently, a regulator may interpret that a trained model violates privacy by recognizing people. In such cases, we must ensure compliance by removing any trace of human patterns from the model while retaining features of other relevant objects in the scene.

Retraining ML models from scratch to accommodate deletions is impractical due to the extensive computational resources required.
As a solution to this challenge, the emerging field of Machine Unlearning (MU) has been proposed to enable models to "forget" specific data points or features without full retraining~\cite{xu2024machine}. This approach efficiently removes sensitive or irrelevant data, addressing privacy concerns. MU not only aids in meeting regulatory requirements but also enhances AI systems' trustworthiness and transparency by reducing biases and preventing adversarial attacks.
The effectiveness and reliability of MU raise several questions, particularly regarding how well patterns are unlearned. While accuracy is a common metric for evaluating ML models, it may not fully capture this nuance, leaving concerns about whether unwanted patterns are genuinely removed or merely re-mapped through different layers of the model.
Explainable AI (XAI) provides insights into ML models' decision-making processes, helping technical end-users to identify and debug errors.
A significant portion of XAI research focuses on attribution-based explainability, which attributes model predictions to specific input features~\cite{colin2022cannot}. These methods provide insights into how much each feature contributes to a model’s prediction. While commonly used to understand and improve model predictions, they are less utilized for verifying MU processes.
 This gap in utilization highlights the need for a deeper examination of MU’s impact on model behavior. By leveraging XAI, we can evaluate whether these methods successfully remove sensitive information and biases or inadvertently introduce new issues. Fig. 1 exemplifies this process in the challenging and real-life test case of harbor front monitoring, an environment that requires ongoing data analysis for effective decision-making. Handling potentially sensitive data, such as images containing people, makes this a prime example of where GDPR compliance is potentially essential. For instance, harbor front monitoring can involve identifying and tracking people for safety and operational purposes, which can include soft biometric data that raises privacy concerns. XAI helps to verify if MU can effectively remove potentially sensitive information from the model’s predictions while retaining relevant object features. The heatmaps generated by XAI show how the trained model captures human patterns, which the unlearned model effectively ignores, ensuring adherence to potentially strict privacy regulations while maintaining accuracy, fairness, and interpretability. This understanding is crucial for validating the model’s performance after unlearning specific data and can potentially highlight the synergy between XAI and MU in building AI systems that are both effective and aligned with ethical standards and societal expectations.

Hence, our research question is:
\begin{itemize}
    \item \textbf{RQ: \textit{Can attribution-based Explainable AI (XAI) be used to verify the effectiveness of unlearning?}}
\end{itemize}
In this paper, we conduct a detailed study on how attribution-based XAI can improve the application of MU techniques by observing and quantifying the shift in local feature importance.  
We explore various MU approaches, including data relabeling and model perturbation, discussing their challenges and potential benefits.
We make use of XAI to define novel metrics and measure, both qualitatively and quantitatively, the impact that MU has on the model's ability to localize appropriate features. Our contributions are threefold:
\begin{itemize}
    \item We demonstrate the novel application of attribution-based XAI methods to verify the effectiveness of MU techniques. By generating saliency maps, we can visualize and qualitatively assess where the models focus post-unlearning, ensuring that sensitive information has been successfully removed.
    \item  We go beyond accuracy measures by introducing two new  XAI-based metrics, Heatmap Coverage (HC) and Attention Shift (AS), to quantify the effectiveness of MU methods. HC measures the spatial correctness of the predicted heatmaps, while AS quantifies the shift in attention to relevant areas, providing a comprehensive assessment of the unlearning process. 
    \item  Our study compares models that have been retrained from scratch with those that have undergone MU processes. We highlight how MU methods, evaluated through XAI techniques,  can efficiently unlearn undesired patterns and potentially offering a more practical alternative to complete model retraining.
\end{itemize}
To the best of our knowledge, this is the first time that XAI is leveraged to verify and inspect the impact of MU techniques.
\section{Related Work}
As AI and ML continue to dominate various domains, ensuring data privacy and model transparency has become crucial. This has led to growing interest in MU techniques, that remove sensitive data without full retraining, complying with GDPR.
Combined with XAI, which promotes transparency in AI decision-making, these approaches can be key to building trustworthy AI systems.
While XAI and MU are active independent fields of study, their combined effects in sensitive applications, such as security monitoring, are to be explored. This integration is vital for predictive analytics and decision-making, where attribution-based XAI methods can ensure data privacy, model interoperability, and verify the efficacy of MU methods. In the following subsections, we review the current research in MU and attribution-based XAI, discussing major techniques and their implications for enhancing privacy and transparency in AI applications.

\subsection{Machine Unlearning}
MU methods can be broadly categorized into three subcategories: \textit{model-agnostic}, \textit{model-intrinsic}, and \textit{data-driven} approaches~ \cite{nguyen2022survey}.\\
\textbf{Model-agnostic} methods apply to various learning models and provide a solid foundation for addressing data deletion and privacy concerns. Several approaches fall into this category. \textit{Certified Removal} formalize the definition of MU and provide methods with theoretical guarantees~\cite{guo2023certifieddataremovalmachine}. Typically they use noise to mask small changes from gradient updates, only suitable for certain convex losses. \textit{Decremental Learning} reduces training load by removing redundant samples or pruning irrelevant nodes~\cite{NEURIPS2023_a204aa68}. \\
\textbf{Model-intrinsic} methods are tailored for specific model types. Approaches for \textit{linear models}, \textit{tree-based models}, and \textit{Bayesian models} involve influence functions, robust split decisions, and optimizing posterior distributions. These are effective but often struggle with scalability and efficiency for complex models. \textit{Deep neural networks} (DNNs), require specialized methods like certified removal mechanisms for layers with convex activation functions or influence functions together with noise injection, and tracking gradient updates on nodes~\cite{9577384}.\\
\textbf{Data-driven} approaches to modify or track data to efficiently remove the influence of specific data samples from the model. For example, SISA technique~\cite{9519428} leverages data partitioning and model ensembling. While effective for small tasks, they do not generalize well with larger models and dataset complexity. Other methods involve data augmentation by adding noise or relabelling data~\cite{li2023randomrelabelingefficientmachine}. Despite potentially achieving good accuracy, these methods lack guarantees of removal, and are not feasible for many practical scenarios.\\
Despite the variety of the proposed methods, they lack generalization to complex models and losses, only handle simple removal tasks, and do not provide a general benchmark to properly compare and verify unlearning efficacy.\\
Recently, the first MU competition at NeurIPS 2023~\cite{triantafillou2024makingprogressunlearningfindings} offered a valuable platform for benchmarking unlearning algorithms. The evaluation focused on the quality of forgetting while also considering model utility, providing a dual perspective for assessing the effectiveness of these methods. In our proof of concept, we compare fine-tuning on relabeled data with decremental learning and noise injection for DNNs, which were top-performing approaches in the challenge. They proposed using an F-score as a metric for evaluating forgetting quality, requiring training a model from scratch as a baseline and running several attacks on it, which is resource-intensive and often unfeasible. Additionally, F-score is primarily defined for measuring the removal of data points in classification tasks. Alternative accuracy and attack-based metrics also lead to similar problems \cite{nguyen2022survey}.
Therefore, there is a pressing need for an interpretable, efficient, and versatile verification mechanism that can assess forgetting quality in classification tasks and beyond. We claim that XAI provides the necessary tools to address this issue.

\subsection{XAI}
As neural networks have expanded in size and complexity, ensuring their interpretability has become a major challenge, making explainability increasingly important. Attribution-based Explainable AI (XAI) methods aim to help users understand the predictions of "black box" models by assigning importance to input features. These methods are typically categorized into \textit{Class Activation Map (CAM)-based methods}, which include gradient-based and gradient-free approaches, \textit{perturbation-based methods}, and \textit{contrastive methods}.\\
\textbf{CAM-based} methods utilize feature maps from the final convolutional layers to determine which regions in an input image are responsible for the model’s predictions. These methods can be further broken down into two subcategories: 1) \textit{Gradient-based} methods compute gradients of the model’s output with respect to the input features to generate saliency maps, highlighting each feature's contribution to the final prediction. Examples are Grad-CAM~\cite{selvaraju2020grad}, Grad-CAM++~ \cite{chattopadhay2018grad}, Layer-CAM~\cite{jiang2021layercam}, and Layer-wise Relevance Propagation (LRP)~\cite{bach2015pixel} fall into this category. These methods require access to model parameters and may struggle with negative contributions due to gradient saturation. 2) \textit{Gradient-free methods} do not rely on gradient information, which makes them advantageous for models in post-deployment settings. Examples include Score-CAM~\cite{wang2020score} and SIDU (Similarity Difference and Uniqueness)~\cite{muddamsetty2022visual}. \\
\textbf{Perturbation-based} methods systematically modify the input features and observe the resulting changes in the model’s output. Techniques like LIME~ \cite{ribeiro2016should} and RISE~\cite{petsiuk2018rise} approximate a model locally or use random perturbations to generate saliency maps, suitable for black-box scenarios. However, they are computationally demanding and limited in applicability.\\
\textbf{Contrastive methods }explain why certain predictions are made and others are not, essential for high-stakes domains like healthcare and autonomous driving~ \cite{abhishek2211attribution}. Techniques like Contrastive Explanation Method (CEM)~\cite{dhurandhar2018explanations} and CDeepEx~\cite{feghahati2020cdeepex} identify features relevant to both positive and negative predictions, offering deep insights but often requiring intensive computation due to complex optimization.\\
While each method category provides valuable insights, many visual explanation techniques face challenges beyond computational complexity, particularly in generating saliency heatmaps that precisely identify and localize the regions responsible for the model’s predictions. Gradient-based methods rely on gradient computations to determine feature importance, which can be problematic for input regions that significantly influence the model’s prediction but yield zero gradients. Perturbation-based methods like LIME introduce randomness through their sampling strategies, leading to variability in the explanations generated.
To address these challenges, we employ SIDU method (see Sec.\ref{SIDU}), a gradient-free and computationally efficient algorithm, which has
demonstrated success in both image~\cite{muddamsetty2022visual} and text domains~\cite{jahromi2024sidu}, particularly in sensitive applications, by effectively highlighting crucial contextual information relevant to the model's predictions. 
\section{Dataset}
The dataset utilized in this study extends the Long-Term Thermal-Imaging Drift (LTD) dataset~\cite{nikolov2021seasons}, covering a period of 188 days from May 14, 2020, to April 30, 2021. 
Due to the dataset being recorded in a public space, the privacy of the individuals in the context is paramount. Thus a request for MU of specific patterns is a real possibility. 


The dataset consists of 1.069.428 images with four annotated classes; Human, Bicycle, Motorcycle and Vehicle. Due to the long-term nature of the dataset captures a wide range of visual appearances, environmental conditions and times of day, emphasizing the dynamic challenges of real-world applications.

The data is divided with a temporally uniform distribution between train, test and validation sets. All frames belonging to a specific video clip, are grouped to prevent data leakage between splits.
Due to the elevated camera position and low resolution of the camera, objects often appear small in the footage, complicating object-centric vision tasks. 
These challenges are further exacerbated by the class imbalance of the dataset caused by the typical behavior in this particular public space. 
Likewise, the frequency of objects in a given image follows a Poisson-like distribution with the most common samples containing 1-4 objects. 
We address this imbalance by sub-sampling the top 70th percentile samples while up-sampling the bottom 10th percentile. 
For validation and testing, however, the distributions remain uniformly sampled across the entire dataset to evaluate real-world performance.

\section{Proposed method}
Typically, advances in MU have been tailored to simple classification tasks, with fewer approaches addressing real-case scenarios involving complex data. Furthermore, the success of these methods is often evaluated based on the performance of the retained data, while the verification of the unlearning method's efficacy is frequently overlooked. To advance the complexity of typical MU tasks and better reflect real-world scenarios, we define a challenging regression problem where the model must count objects of interest from the LTD dataset and unlearn counting humans. We propose leveraging attribute-based XAI to evaluate how much MU succeeds qualitatively and quantitavely in forgetting learned patterns (as shown in Fig. 2). Particularly, by means of SIDU that generates a heatmap, based on the impact of different channels (i.e. abstract patterns) in the latent space of the model. By comparing the heatmaps from a model naively trained from scratch without counting humans, and those of models exposed to MU methods, we inspect and evaluate the efficacy of different unlearning methods on the task of object counting.  

\begin{figure}
    \centering
    \includegraphics[width=\textwidth]{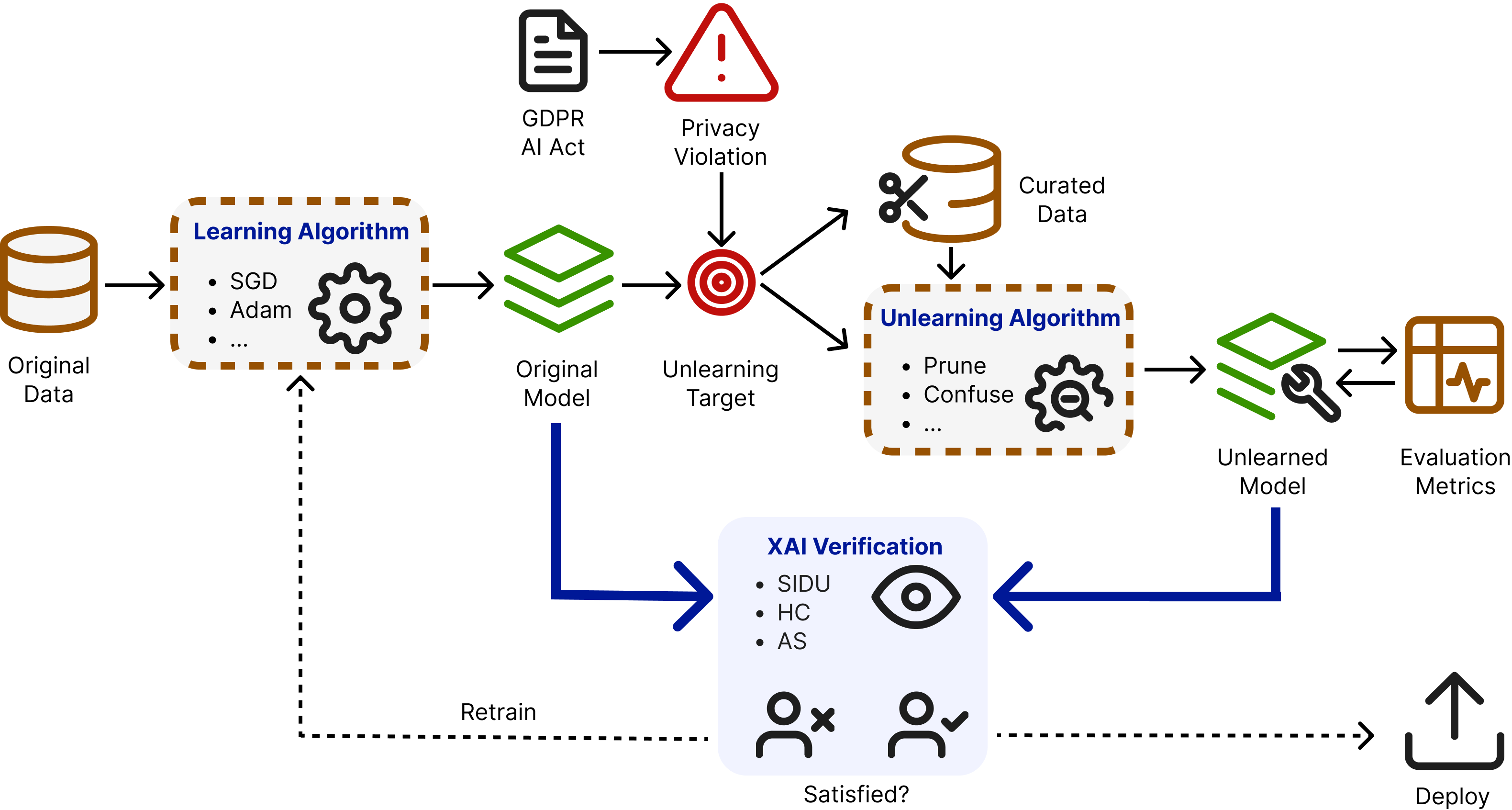}
    \caption{Our framework uses explainability to verify unlearning. Upon a removal request, the model undergoes unlearning to exclude people from the counting task, with SIDU verifying the unlearning success.}
    \label{fig:mu-xai-framework}
\end{figure}
\vspace{-1cm}
\subsection{\textbf{Problem statement}}

Let $D$ be the LTD dataset, so that if $(\textbf{x},y)\in D$, then $\textbf{x}$ is a vector representing a thermal image and the label $y$ is a scalar that counts the objects of interest in such image. Originally, the dataset contains four classes: human, bicycle, vehicle, and motorcycle; and tracks them over a video stream. We obtain $D$ by summing the total number of instances that appear at any frame, that is, for each thermal image $\textbf{x}$ we have $y = y_h + y_b + y_v + y_m$.

Inferring the number of relevant objects in the scene is important for tracking overall behavior. Typically, a model $f$ is trained on dataset $D$ to perform this task. However, monitoring peoples' behavior is highly sensitive and subject to GDPR legislation.
The GDPR strictly regulates personal data, including images with people. Thus, a regulator may deem model $f$ unwanted for counting individuals, requiring the removal of people from the model while still counting other relevant objects in the scene.
In this situation we can take two different paths, either train a new model removing $y_h$ from $D$, or modify $f$ through an unlearning method to count only the remaining classes. After the unlearning process, we leverage SIDU as a verification tool to certify that $f$  has successfully "forgotten" to be influenced by human-like patterns in the attribute space.

\vspace{-.3cm}
\subsection{\textbf{Original Model Training}}

For each element in $D$ we train a model $f$ to infer the amount of objects of interest in our scene. This is a typical regression problem, that can be solved by minimizing a loss function between the predictions and the ground truth labels. 
Formally, if $(\textbf{x}, y)\in D$, then $f_\mathbf{\theta}(\textbf{x}) = \hat{y}$ is the prediction of the model, and $\mathbf{\theta} \in \mathbb{R}^d$ are the parameters of the model. The training objective consists of minimizing the problem
$$\arg\min_\mathbf{\theta} \mathcal{L}(f_\mathbf{\theta}(\textbf{x}), y)$$
where $\mathcal{L}$ is a loss function. We are using the Mean Squared Error (MSE) for this use-case. Once trained, if $\mathbf{\theta}^*$ are the optimal parameters of the model, we denote $f_o = f_{\mathbf{\theta}^*}$ the Original model.

\subsubsection{Evaluation}
To evaluate the performance of the trained model, we first measure it with common regression metrics, such as RMSE and MAE. Then, we use SIDU to generate the heatmaps from the images that reveal the patterns the model focuses on. While RMSE and MAE provide quantitative metrics, SIDU offers a qualitative assessment of the model's performance by visualizing its attention to different features.
\vspace{-.3cm}
\subsection{\textbf{Unlearning Methodology}}\label{subsec:unlearning_methodology}



In our proof-of-concept, we want to unlearn the ability to count humans and that information must also be deleted from $D$. Since the LTD dataset is separated by classes and we know per class instances, we remove the counting people ability from the dataset, by updating the label: $y' = y_b + y_v + y_m$. Thus, we obtain the curated dataset $D' = \{(\textbf{x}, y') \; | \; (\textbf{x}, y) \in D\}$, see Fig. 3. This can be generalized to other counting problems that use object tracking datasets, which are also typically label class-wise.

\begin{figure}[ht]
    \centering
    \includegraphics[width=\textwidth]{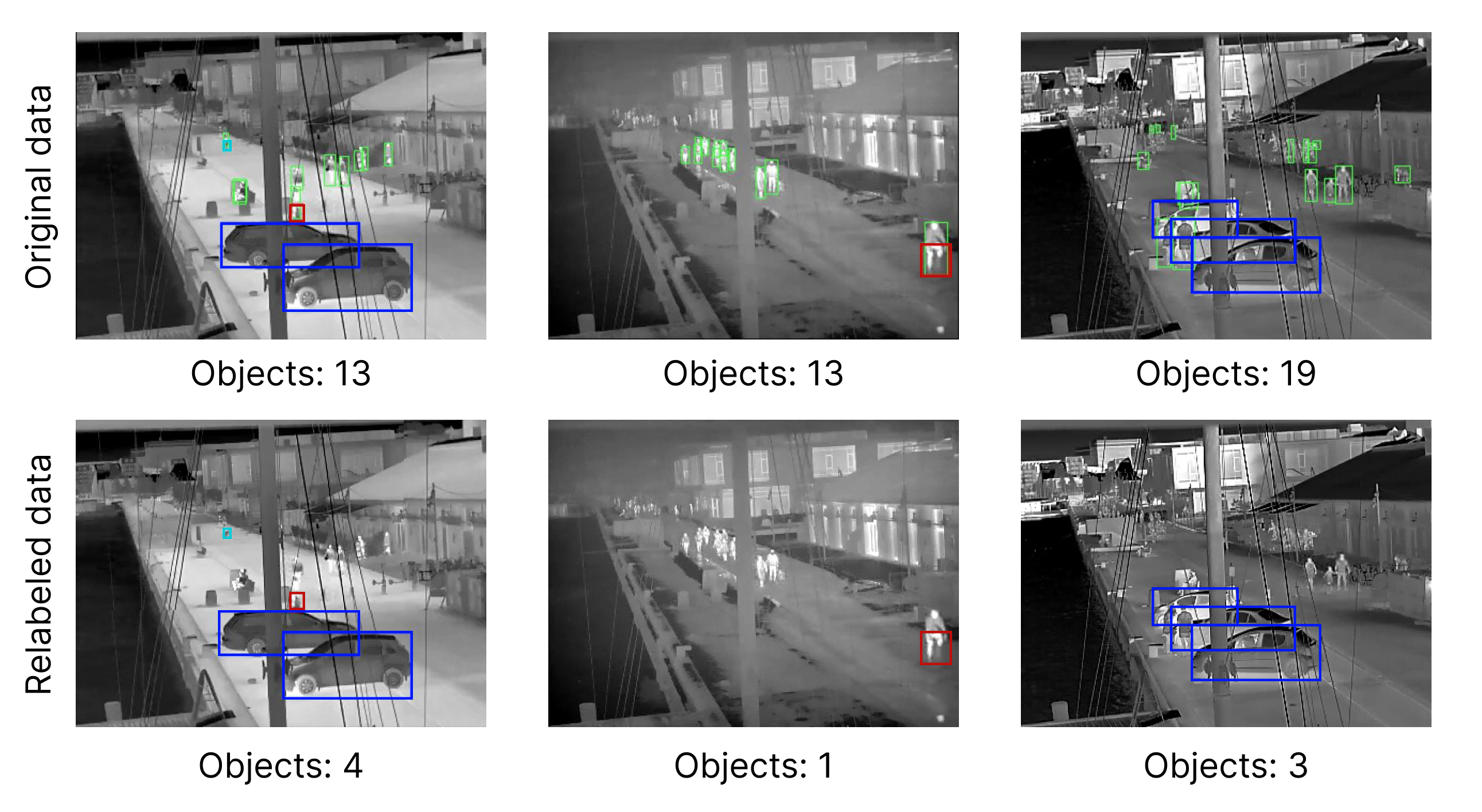}
    \caption{Examples from Original dataset and Relabeled dataset. The color of a bounding box denotes the object class, with green, red, blue and cyan denoting human, bicycle, vehicle and motorcycle respectively.}
    \label{fig:modified-data}
\end{figure}

    

The naive approach to address this issue consists on instantiating a new model $f_b$ and solve the regression problem as before but over the curated dataset $D'$. This ensures that the outcome model do not track human attributes. However, going through all the training process again is not efficient and not feasible in a practical case. An unlearned model starts from $f$ and modify it to achieve a similar behaviour as $f_b$. We denote the unlearned model by $f_u = U(f,D')$, where $U$ is an unlearning method, and call $f_b$ the Baseline model. While our approach involves relabeling as part of the unlearning process, this step alone is insufficient to fully address the unlearning challenge. Relabeling adjusts the task to avoid counting humans, but the model may still retain internal representations of human features. To fully remove these learned features from the model, additional techniques such as decremental learning and noise injection are required to unlearn human-related patterns effectively. This ensures that the unlearned model no longer relies on human features in its internal layers, addressing the limitations of fine-tuning or simple relabeling alone. Inspired by~\cite{triantafillou2024makingprogressunlearningfindings}, we compare the retrained-from-scratch Baseline with four different models:

\begin{itemize}
    \item \textbf{Fine-tune:} Simple approach that consists of fine-tuning $f$ with the relabelled data $D'$, i.e. minimizing MSE starting with $f$ iterating over $D'$. We denote the Fine-tune model by $f_{u_f}$. For small tasks, this has been proven to be enough due to catastrophic forgetting~\cite{Kemker_McClure_Abitino_Hayes_Kanan_2018}.
    \item \textbf{Prune and Reinit}: This method achieved the 4th position within the NeurIPS 2023 MU challenge and the highest forgetting score in the post-challenge bench-marking~\cite{triantafillou2024makingprogressunlearningfindings}. It consists of pruning ($f_{u_p}$) or reinitialize ($f_{u_r}$) the 95\% of the convolutional and fully-connected layer weights with the lowest L1 norm. The model then performs fine-tuning on the retain set using a combination of loss function and regularization term between the entropy of the original model and the entropy of the unlearned model.
    \item \textbf{Confuse:} Inspired from Seif method \cite{triantafillou2024makingprogressunlearningfindings}, which achieved the 3rd position, the Confuse method adds Gaussian noise to convolutional weights. That is, if $\theta^*_{\text{conv}} \subset \theta^*$ are the convolutional parameters of $f_o$, then we obtain the confused model $f_{u_c}$ by cloning the original model but setting the convolutional weights to be $\theta \sim \mathcal{N}(\mu = \theta^*_{\text{conv}}, \sigma^2 \cdot I)$. The hyper-parameter $\sigma$ controls the scale of the added noise. 
\end{itemize}


\subsubsection{Adapting Challenge Methods}
Note that Prune, Reinit and Confuse are methods directly inspired from the top solutions from the NeurIPS 2023 MU Challenge, which consisted of performing unlearning in a classification task. Since in our use-case we are dealing with a regression problem, these methods have been adapted to the regression needs. Prune and Reinit skip the entropy regularization term, since output of $f_{u_p}$ and $f_{u_r}$ is a single unbounded scalar instead of a vector of logits. Confuse differs from Seif by skipping a weighting loss term that aimed to balance loss according to the classes within a mini-batch. As in Fine-tune, the models $f_{u_p}$, $f_{u_r}$ and $f_{u_c}$ are obtained by applying their respective unlearning methods and minimizing the MSE over $D'$.

\subsection{\textbf{XAI model: Similarity Difference and Uniqueness (SIDU)}}\label{SIDU}

SIDU estimates pixel-wise importance by extracting feature maps from the last convolutional layer of a deep CNN and creating masks based on similarity differences (SD) and uniqueness (U) metrics (See Eq. 3 and Eq. 4 in~\cite{muddamsetty2022visual}). SD measures how altering a feature impacts the model’s output, while Uniqueness assesses the distinctiveness of each feature. The combination of these masks produces a comprehensive heatmap that efficiently highlights influential elements in the model’s decision-making. This method addresses the limitations of existing techniques by more accurately localizing the salient regions that contribute to the model’s predictions. Our experiments have shown that SIDU not only performs well in identifying these regions but also enhances trust in the model's decisions, particularly in sensitive domains like medical imaging. Additionally, SIDU has demonstrated robustness against adversarial attacks, making it a reliable and effective tool for visual explanations in AI.
Typically, approaches to unlearning involve retraining or fine-tuning the model, assuming that performance degradation on the undesired patterns indicates successful forgetting. However, these methods often overlook the need to qualitatively verify the unlearning process. SIDU addresses this gap by generating detailed attribution heatmaps that visualize the contribution of different input regions to the model’s predictions. 
For instance, in the context of a monitoring scenario, where the model is supposed to "forget" human-related patterns due to privacy regulations, the SIDU heatmaps should show a significant reduction in the focus on human features post-unlearning. This visualization provides a clear, intuitive way to ensure that the model no longer relies on the undesired patterns, thereby validating the success of the unlearning process in maintaining compliance with privacy standards.

\subsection{Metrics}\label{subsec:metrics}
To quantify the performance of the proposed network, we employ two well-established regression metrics, MAE and RMSE, to compare model predictions and labels. Evaluating how effectively an attribution method like SIDU highlights relevant parts of an image remains an open problem. To this end, we propose a novel metrics Heatmap Coverage (HC). Additionally, we define Attention Shift (AS) to measure the attention drift between $f_o$ and $f_{u_i}$.
\begin{equation}
    \text{HC}(f_i) = \frac{1}{N}\sum_{j=1}^{N} \frac{\text{sum}(\textbf{H}_i^j \odot  \textbf{M}_i^j)}{\text{sum}(\textbf{H}_i^j)}
    \label{eq:hmcvr}
\end{equation}
As SIDU produces heatmaps that rank the relative importance of areas within the image, HC quantifies the spatial correctness of the predicted heatmaps by computing the average of a weighted element-wise overlap between predicted SIDU heatmap $\textbf{H}$ and its corresponding region of interest mask $\textbf{M}$, see \Cref{eq:hmcvr}.
Where $\text{sum}(\cdot)$ is the summation of all the elements of the matrix. $\textbf{H}$ and $\textbf{M}$ are matrices of dimension [$\text{Image}_{\text{width}}$, $\text{Image}_{\text{height}}$].
Matrix $\textbf{M}$ is generated by using the object detection bounding box annotations, where every pixel that falls inside the bounding box of an object of interest is given the value 1, and 0 otherwise.
\begin{equation}
    \text{AS}(f_{u_i},f_o) = \frac{1}{N}\sum_{j=1}^{N} \text{std}(\textbf{H}_{u_i}^{j} - \textbf{H}_{o}^{j})
    \label{eq:attshift}
\end{equation}
When comparing heatmaps generated by the original model with those from the unlearned model, we measure the change in relative importance by computing the standard deviation of the difference between heatmaps, see \Cref{eq:attshift}. This allows us to quantify the shift in attention to relevant areas of the images.

Where $u_i$ indicates the unlearned model, $o$ the Original model, and $\text{std}(\cdot)$ computes the standard deviation from all elements of the matrix.

\section{Results}\label{sec:results}
To establish a baseline we train two models (Original and Retrain) which are randomly initialized and trained for $10$ epochs. As unlearning assumes a baseline starting point, MU models are initialized from the baseline and train for $3$ epochs. Each model uses a batch size of $50$ and is optimized with the SGD optimizer at a learning rate of $5e-4$. These models are then evaluated qualitatively and quantitatively  with the metrics described in \Cref{subsec:metrics}.

\subsection{Quantitative results}
As can be seen in \Cref{tab:performance_table} all unlearning methods yield similar quantitative results, but they fall short of baseline Retrain. This is expected, as the MU models' training process is 3 times shorter. However, a closer examination of r-HC (Heatmap Coverage for the retaining classes) indicates that MU methods concentrate heatmap intensity more closely on the objects we aim to retain. This behavior suggests that an unlearned model can better retrieve object patterns after unlearning human ones. This improvement could be due to the high-class correlation within the dataset; for instance, a bike is often associated with a human riding it.
\begin{table}[]
    \centering
    \caption{Overview of the different metrics for the models. The bolded values represent the best scores for each respective metric across the unlearning methods.}
    \begin{tabular}{|l|c|c|c|c|c|}
        \hline
        \textbf{Model} & $\downarrow$ MAE & $\downarrow$ RMSE & $\uparrow$ r-HC & $\downarrow$ h-HC & $\uparrow$ AS\\\hline
        Original    & 2.0430 & 3.3170 & 3.903e-3 & 3.977e-3 & - \\            
        Retrain     & 0.5531 & 0.8877 & 3.299e-3 & 1.223e-3 & 2.085e-5\\\hline 
        Finetune    & 0.6112 & 0.9522 & 6.827e-3 & 2.802e-3 & 1.426e-5\\       
        Prune    & 0.6128 & 0.9526 & 6.434e-3 & 2.511e-3 & 1.391e-5\\       
        Reinit   & \textbf{0.5693} & \textbf{0.9090} & 5.464e-3 & 2.214e-3 & 1.486e-5\\       
        Confuse     & 0.5866 & 0.9327 & \textbf{7.632e-3} & \textbf{1.989e-3} & \textbf{1.492e-5}\\\hline       
    \end{tabular}\label{tab:performance_table}
\end{table}
Furthermore, Confuse noticeably better localize the attention around objects of interest compared to the other models. Conversely, h-HC (Heatmap Coverage for human class) indicates that heatmaps from the Retrain model are less weighted around human bounding boxes, suggesting that the model's attention is dispersed elsewhere in the image. Interestingly, Prune, Reinit, and Confuse exhibit lower h-HC values than Finetune, underscoring that purposefully designed MU strategies can effectively reduce the attention models give to human attributes in the latent space, thereby demonstrating the effectiveness of unlearning.
When considering AS, Confuse has shifted more attention compared to the original heatmap, which is also reflected in r-HC and h-HC. As expected, none of the models have shifted the attention as much as Retrain, indicating that some attention to human patterns still remains.
\vspace{-.3cm}
\subsection{Qualitative results}
Each heatmap in Fig. 4 visualizes the regions of the image that the model focuses on.
The heatmaps in (c) Original and (d) Retrain show the initial model's focus areas before any unlearning techniques were applied. Notably, these heatmaps include significant attention on regions associated with human presence.
\begin{figure}[ht]
    \centering
    \includegraphics[width=1\textwidth]{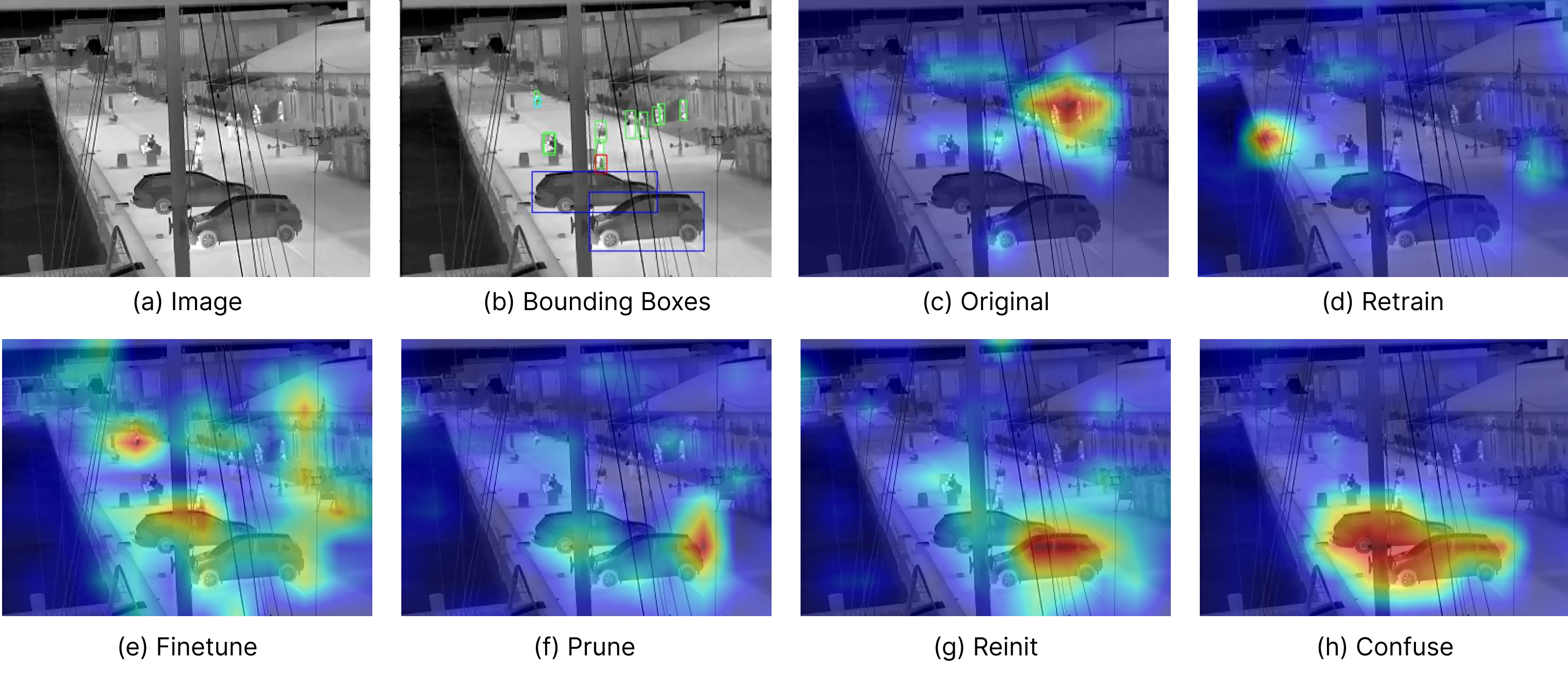}
    \caption{This figure shows heatmaps generated by SIDU for different model configurations. Where importance is visualized from least (blue) to most (red).}
    \label{fig:qual_examples}
\end{figure}
\newpage
\begin{figure}[ht]
    \centering
    \includegraphics[width=\textwidth]{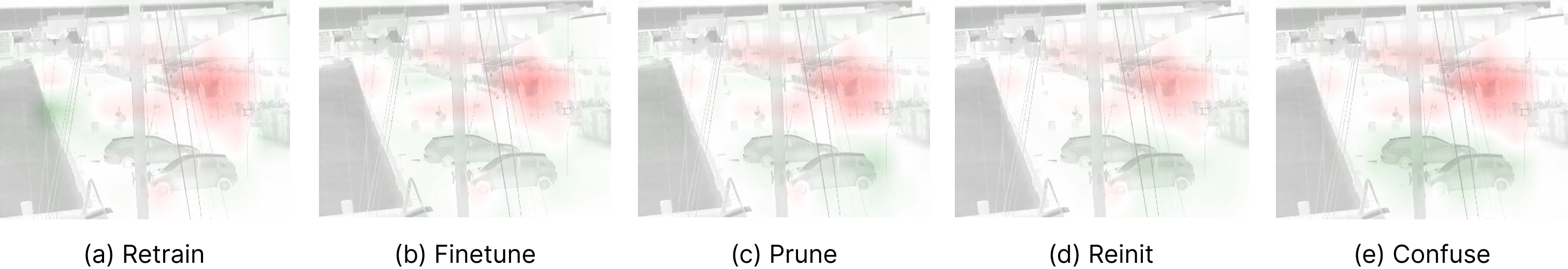}
    \caption{This figure illustrates the difference between the original heatmap and the unlearned. Green indicates increased attention, red indicates decreased attention and white areas minimal change.}
    \label{fig:att_shift}
\end{figure}
\vspace{-.3cm}
In contrast, heatmaps from the models processed with MU techniques—(e) Finetune, (f) Prune, (g) Reinit, and (h) Confuse—demonstrate varying levels of reduced attention on human-related regions. Finetune (e) and Prune (f) methods show a noticeable shift in attention, focusing more accurately on non-human objects, such as bicycles and vehicles, indicating the removal of human-related features. The Reinit (g) and Confuse (h) methods also exhibit an effective reduction in human feature importance, with the Confuse method showing particularly dispersed attention away from human regions, thereby validating the effectiveness of these MU methods in eliminating human features from the model's focus.
These trends are further highlighted in Fig. 5. It can be seen that all methods that are not trained to count humans, have shifted their attention to the regions related to the remaining objects.
\vspace{-.3cm}
\subsection{Discussion}
As outlined in this section the difference in the unlearning approach in terms of standard evaluation metrics (MAE \& RMSE) are minor. However, MU techniques can help focus the model's attention to an extent that surpasses straightforward fine-tuning. This not only underlines the importance of employing MU techniques to accelerate the removal of patterns by reducing the computational burden but also highlights that doing so will allow the model to focus more narrowly on the remaining content. Furthermore, by visually comparing these heatmaps (as shown in Fig. 4), we can qualitatively assess the success of each unlearning method. The reduction in attention on human-related features in the Prune and Confuse methods, in particular, highlights their effectiveness in ensuring compliance with privacy regulations while maintaining the model's ability to focus on other relevant objects in the scene. One key consideration is the assumption that the XAI methods used for heatmap generation accurately reflect important regions in the image. While XAI methods offer valuable insights into model decision-making, the absence of a consensus on the efficacy of XAI techniques as well as metrics for evaluating model explanations adds a layer of uncertainty which should be acknowledged. Despite these challenges, this visual inspection underscores the utility of XAI as a tool for verifying the efficacy of MU techniques in sensitive applications. 
Effectively the purpose of MU to adhere to the "right to be forgotten", would be to ensure that the undesired pattern is disregarded and does not play a significant role in the system's decision-making. This notion of attribution-based evaluation of efficacy lends itself naturally to be addressed with the use of attribution-based explanations. 
\vspace{-.3cm}

\section{Conclusion}
In the evolving landscape of AI systems, privacy and transparency have become crucial due to AI's increasing complexity and impact on daily life. 
MU emerged to help address the privacy need, it necessitates a verification method to properly evaluate whether it has the intended effect.
As XAI provides insights on the decision-making processes of ML models, we claim it is a crucial tool to address MU challenges.
Our findings show that using to attribution-based XAI, unlearned models can better concentrate heatmap intensity on the desired target objects when compared to completely retraining the model. 
While fine-tuning the model shows a better concentration on retaining classes than some MU approaches, XAI reveals that traces of unlearned patterns still linger, which indicates inefficient unlearning.
Our novel approach could further expand to increasingly complex systems and could be applied with other attribution-based XAI methods, to provide additional insight into existing and future systems. 
The presented methodology demonstrates significant potential for XAI in evaluating MU methods. Additionally, after leveraging different verification scenarios, the next step can be joint end-to-end optimization to ensure that XAI features are considered to directly target undesired attributes at training time, thus enhancing the robustness and applicability of MU frameworks.

\begin{credits}
\subsubsection{\ackname} This work was conducted as a part of the Responsible AI for Value Creation project (REPAI), grant no. 83648813, by the Grundfos Foundation, Contestable AI, grant no. 10.46540/2027-00140B by the Independent Research Fund Denmark, the Spanish project PID2022-136436NB-I00, Milestone Research Program at AAU, and the ICREA Academia programme.

\subsubsection{\discintname}
The authors declare that they have no known competing financial interests or personal relationships that could have appeared to influence the work reported in this paper.
\end{credits}
%
%
%
\bibliographystyle{splncs04}
\bibliography{references}
%






\end{document}